# Co-word Analysis using the Chinese Character Set

Loet Leydesdorff [a] & Ping Zhou [b,c]

**Abstract**

Until recently, Chinese texts could not be studied using co-word analysis because the words are not separated by spaces in Chinese (and Japanese). A word can be composed of one or more characters. The online availability of programs that separate Chinese texts makes it possible to analyze them using semantic maps. Chinese characters contain not only information, but also meaning. This may enhance the readability of semantic maps. In this study, we analyze 58 words which occur ten or more times in the 1652 journal titles of the *China Scientific and Technical Papers and Citations Database*. The word occurrence matrix is visualized and factor-analyzed.

[a] Amsterdam School of Communications Research (ASCoR), University of Amsterdam, Kloveniersburgwal 48, 1012 CX Amsterdam, The Netherlands; loet@leydesdorff.net; http://www.leydesdorff.net
[b] K.U. Leuven, Steunpunt O&O Indicatoren, Dekenstraat 2, B-3000 Belgium; Ping.Zhou@econ.kuleuven.be
[c] Institute of Scientific and Technical Information of China, 15 Fuxing Road, Beijing, 100038, P. R. China.

**Introduction**

Unlike most languages, Chinese does not use spaces to separate characters into words. Co-word analysis (Van Rijsbergen, 1977; Salton & McGill, 1983; Callon *et al*., 1982 and 1986; Leydesdorff, 1989 and 1997) has therefore been unable to use Chinese texts. Given the increased importance of the Chinese contribution to science (Zhou & Leydesdorff, 2006), the mapping of co-words in this language has been desirable for some time. (In a previous study, Park & Leydesdorff (2004) solved the problem of mapping texts using the Korean character set; a set of freeware programs was brought online for this purpose at http://www.leydesdorff.net/krkwic.)

Recently, Chen (2007) reported a co-word map using the "traditional Chinese" character set based on software developed by the Academica Sinica in Taiwan (at http://ckipsvr.iis.sinica.edu.tw/). Similar software is available in mainland China for "simplified Chinese." In an analysis of Japanese policy documents based on parsing of the various character sets in Japanese with dedicated software, Fujigaki & Nagata (1998, at p. 394, note 6) suggested that word and co-word analysis may be more meaningful when using the Chinese character set because these characters (unlike the phonetic ones)[1] would contain not only information, but also meaning.

In this brief communication, we explore co-word analysis using Chinese characters and compare the results against our background knowledge of using co-word analysis in

---

[1] In addition to the Chinese ("Kanji") characters, Japanese uses two other sets (Hiragana and Katakana) containing phonetic characters.

English (Leydesdorff, 1989 and 1997). For that purpose, we apply one of these tools—中文智能分词available at http://www.hylanda.com/product/fenci/tiyan/index.html—to the list of journal titles contained in the *China Scientific and Technical Papers and Citations Database* (CSTPCD) database.

**Data and methods**

The data consists of the 1652 journal titles contained in the *China Scientific and Technical Papers and Citations Database* (CSTPCD) of the Institute of Scientific and Technical Information in Beijing in 2005. The CSTPCD is a database of Chinese journals organized in a manner comparable to the *Science Citation Index* of Thomson Scientific/ISI (Zhou & Leydesdorff, 2007). Of the 1652 journals only 36 were published under titles in English. Ren (2005) estimated that approximately 5,000 scientific journals are published regularly in the People's Republic of China.

1157 words occur 4509 times in these 1652 titles. Among these words 697 occur only once; the 58 words that occur ten or more times were used in this analysis. Only two words among them ("Chinese" and "Journal") are in English. These 58 words were cross-tabled against the 1652 titles. The asymmetrical occurrence matrix was normalized using the cosine for the normalization of the word vectors (Ahlgren *et al*., 2003; Leydesdorff & Vaughan, 2006). The cosine matrix was taken for the visualization using Pajek, and the data matrix was also factor-analyzed using SPSS.

**Results**

Figure 1 provides the resulting co-word map. The threshold is set at cosine > 0.07 which is the average value for the cosine in the matrix when the cells with zeros are not included.[2] The size of the nodes is proportional to the logarithm of the number of occurrences of the word, and the thickness of the lines is drawn proportionally to the strength of the relationship (as measured by the cosine).

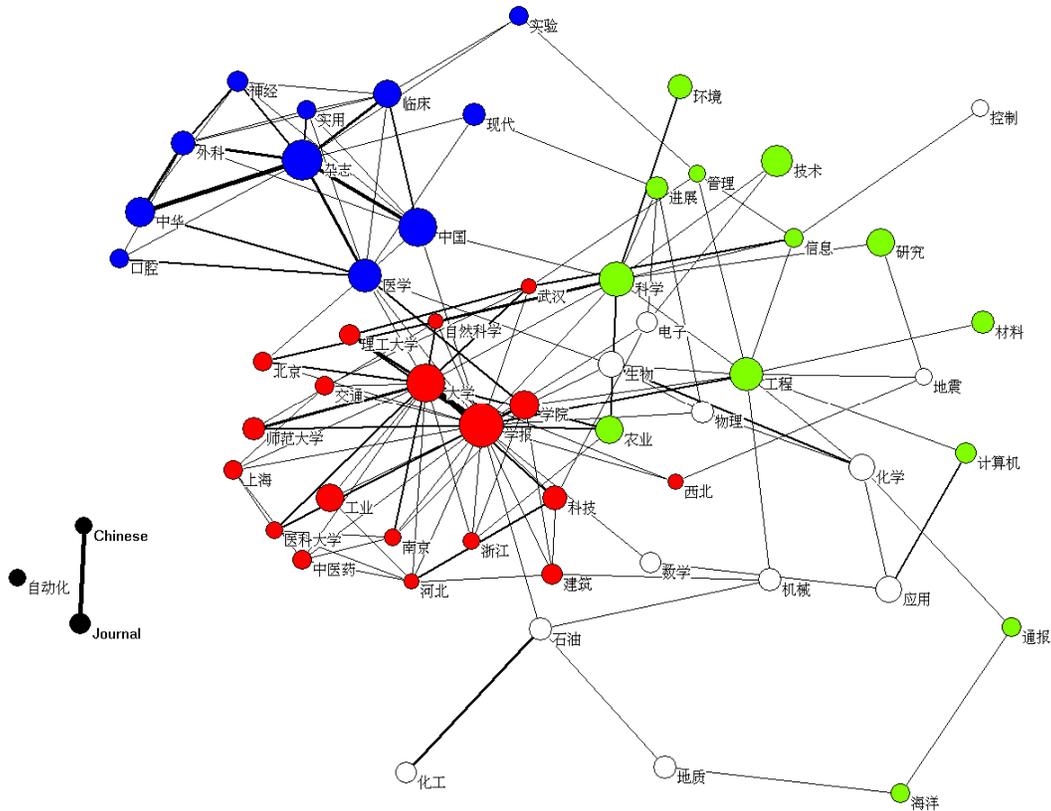

**Figure 1**: Cosine-normalized map of 58 title words occurring in 1652 titles of Chinese journals included in the *CSTPCD* 2005. Red: factor 1; blue: factor 2; black factor 3; green: factor 4; white: no positive factor loadings.

---

[2] The use of the average as a threshold is common in social network analysis. With inclusion of the zeros, the average value for the cosine is 0.019, but this threshold is too low for generating a meaningful visualization.

The two non-Chinese words form a separate group, as was to be expected. They load together on a third factor in the four-factor solution used for coloring Figure 1. The red cluster consists of words which have their highest factor loading on factor one (explaining 4.6% of the variance). This is a collection of words used in the titles of university journals. Titles of Chinese university journals are usually named with a university name followed by "学报" (which means "Journal" or "Transactions"), as for example in the case of the *Transactions of Beijing the Institute of Technology* (北京理工大学学报), where "北京理工大学" is the Chinese name of the institute, and "学报" is added as a suffix to indicate that this is the journal of this institute. Since almost every university or college journal contains the word "学报" (academic journals) in its title, this word has the highest frequency among the nodes colored red. The node labeled as "大学" indicates "university"; "学院" (college) follows with a smaller node size. A number of journals are from technological universities: the Chinese word "工业" means "technology" or "industry."

The second factor (3.3% of the variance) is indicated as a cluster with blue nodes. This cluster can be considered as a representation of journals in the medical sciences. The number of journals with "医学" ("medical science") in their titles is larger than those with "临床" ("clinical science") in their titles, as is indicated by the size of the respective nodes.

In summary, factors one and two correspond to university journals and medical journals, respectively. These are the two main categories of journals in the Chinese set (Jin & Leydesdorff, 2005). Factors three and four (explaining only 2.6 and 2.5% of the variance, respectively) represent title words from journals other than university journals and medical journals. According to one of the referees, factor 3 can be designated as science journals, and factor 4 as technology journals. However, the cosines among these word patterns are not strong enough to form clusters in the semantic map.

**Conclusion and discussion**

Words and co-occurrences of words are not very specific indicators as compared with citations ((Braam *et al*., 1991a and 1991b; Leydesdorff, 1989). In general, the factor structure of the word occurrence matrix is rather flat. Despite low eigenvalues, the first factors can often be provided with a meaningful interpretation (Leydesdorff & Hellsten, 2005). In this case, 27 eigenvectors of the matrix of 58 words as variables have an eigenvalue of one or more, and would therefore explain more than one variable on average. The four-factor solution used above explains only 13.0% of the total variance in the matrix.

Nevertheless, the semantic organization in Figure 1 is meaningful. Perhaps, the factor designation is further supported by specific meaning contained in Chinese characters. Given the indicated option to separate Chinese texts automatically, this hypothesis can be tested by drawing samples of Chinese texts and compare their co-word structures with those in their English translations. Translation software between English and Chinese is increasingly available for the automation of these processes. However, our results did not indicate any differences between using Chinese characters or words written in the Latin alphabet. The patterns (factor loadings, eigenvectors, screeplot, etc.) are similar to the ones known from the co-word analysis of English titles. Thus, the use of Chinese characters themselves or in combination with character sets in other languages seems rather unproblematic.

The use of words and co-words for the visualization of semantic maps remains a powerful technique because it enables us to compare across large document sets in different domains (Callon *et al*., 1986; Leydesdorff & Hellsten, 2005). The application is not confined to titles, but can also be used to study full texts (Leydesdorff & Hellsten, 2006). The availability of computer programs for the separation of character strings into words makes the analysis of both full texts and titles an accessible domain for systematic analysis. We have developed computer programs for this purpose that are freely available for academic usage at http://www.leydesdorff.net/software/chinese/index.htm .

**Acknowledgement**

The authors are grateful to Hylanda (海量信息技术有限公司) for providing the program for word separation used for this research.